\documentclass[letterpaper, 10 pt, conference]{ieeeconf}  
\IEEEoverridecommandlockouts                              

\usepackage{amsmath} 
\usepackage{amssymb}  
\usepackage{graphicx}
\usepackage{hyperref}
\usepackage[backend=biber,
url=false,
maxbibnames=100,
maxcitenames=3,
minnames=1,
style=ieee]{biblatex}
\usepackage{tikz}
\usepackage{adjustbox}

\newcommand\submittedtext{%
	\scriptsize This work has been submitted to the IEEE for possible publication. Copyright may be transferred without notice, after which this version may no longer be accessible.}

\newcommand\submittednotice{%

		\begin{tikzpicture}[remember picture,overlay]
			\node[anchor=south,yshift=0pt] at (current page.south) 	{\fbox{\parbox{\dimexpr0.95\textwidth-\fboxsep-\fboxrule\relax}{\submittedtext}}};
		\end{tikzpicture}%
}

\graphicspath{ {images/} } 

\title{\LARGE \bf
	Precision-Focused Reinforcement Learning Model for\\Robotic Object Pushing
}

\author{Lara Bergmann$^{1}$, David Leins$^{1}$, Robert Haschke$^{1}$, and Klaus Neumann$^{1,2}$
	\thanks{*This work was supported by the Fraunhofer InternaI Programs under Grant No. SME 40-09551}
	\thanks{$^{1}$CITEC, Faculty of Technology, Bielefeld~University, Germany\newline
		{\tt\small \{lara.bergmann, klaus.neumann\}@uni-bielefeld.de}\newline
		{\tt\small \{dleins, rhaschke\}@techfak.uni-bielefeld.de}}%
	\thanks{$^{2}$Fraunhofer IOSB-INA, Lemgo, Germany}%
}

\begin{document}
	\maketitle
	\thispagestyle{empty}
	\pagestyle{empty}
	
	\submittednotice

	\begin{abstract}
		Non-prehensile manipulation, such as pushing objects to a desired target position, is an important skill for robots to assist humans in everyday situations. However, the task is challenging due to the large variety of objects with different and sometimes unknown physical properties, such as shape, size, mass, and friction. This can lead to the object overshooting its target position, requiring fast corrective movements of the robot around the object, especially in cases where objects need to be precisely pushed. In this paper, we improve the state-of-the-art by introducing a new memory-based vision-proprioception RL model to push objects more precisely to target positions using fewer corrective movements.
	\end{abstract}
	
	\section{Introduction}
\begin{figure}[]
		\centering
		\includegraphics[width=0.94\linewidth]{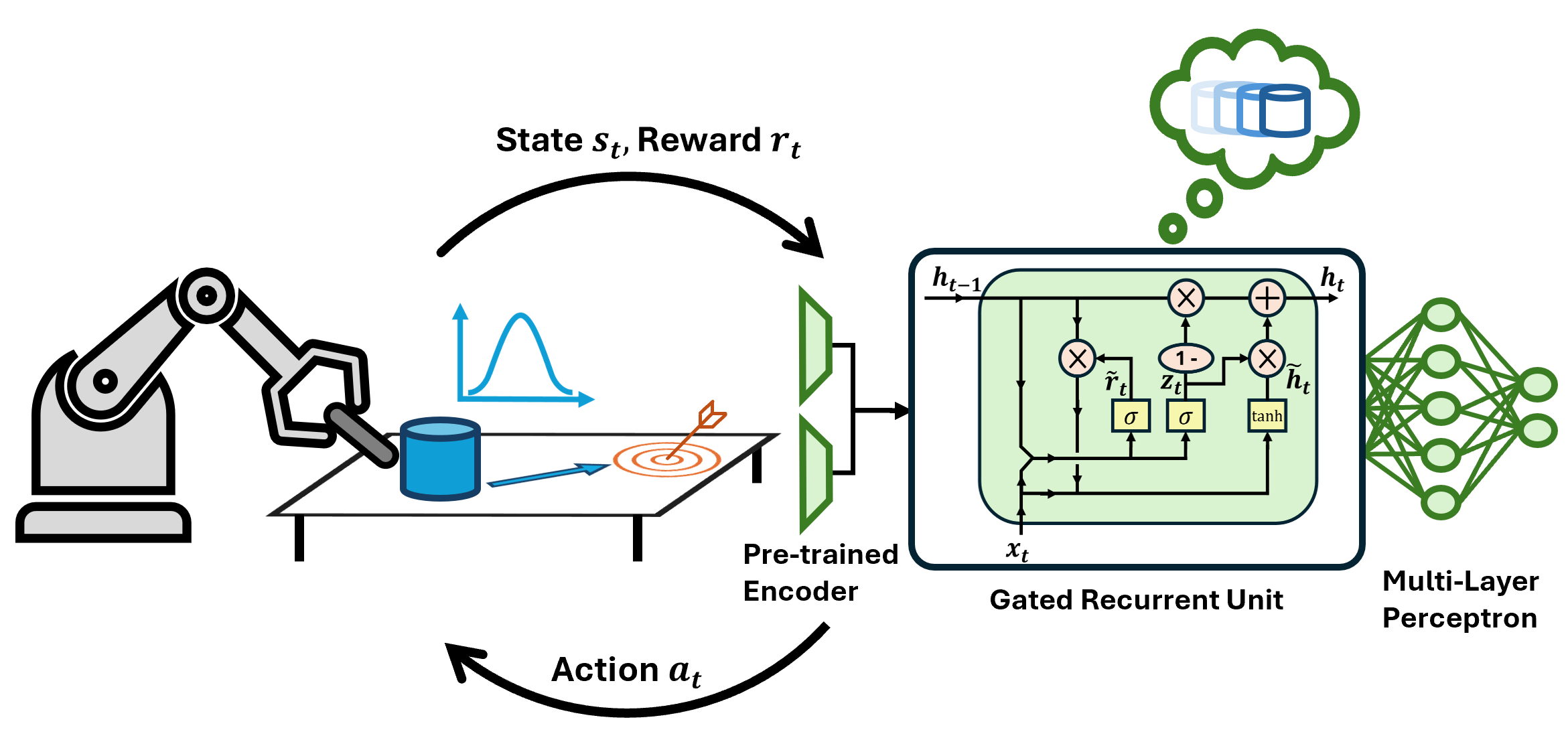}
		\caption{\textbf{Schematic Overview.} We adapt a state-of-the-art RL model to push objects more precisely to target positions by improving the sampling of object parameters and adding a gated recurrent unit to provide the agent with a memory.}
		\label{fig_visual_abstract}
	\end{figure}

	Humans intuitively interact with objects in everyday situations, often without explicitly planning or thinking about how objects will behave. Non-prehensile object manipulation is an important skill for robots that are designed to assist humans. This work focuses on object pushing, a sub class of robotic manipulation that is crucial e.g. for service robots working in a kitchen, but it can also be beneficial for industrial applications. Consider a robot that has to grasp a cup that is placed on a shelf behind other objects. A simple strategy to reach the desired cup is to push the other items aside. Pushing can also be considered as an alternative to pick-and-place, e.g. if objects are too heavy or too large to be grasped. In addition, fragile objects that are likely to be damaged when grasped can be pushed. However, object pushing is demanding for robots due to the large variety of objects that all behave differently depending on their shape, size, mass, and friction. The task becomes even more challenging, considering that not all physical properties, such as mass or friction, are directly observable. These unknown properties can lead to the object overshooting its target position, requiring fast corrective movements of the robot around the object that are particularly difficult to model explicitly, as they require decisions about when a correction is necessary, how to adapt the pushing direction, and how to plan the corresponding movement. Corrections are even more challenging if the approach should generalize to objects with varying physical parameters. The difficulty of modeling such corrective movements is also evident in other recent work. \citeauthor{cong_self-adapting_2020} \cite{cong_self-adapting_2020} report that their model for object pushing based on a recurrent neural network (RNN) and model predictive control (MPC) cannot properly switch pushing sides, i.e. the model is not able to perform corrective movements. Additionally, the authors also train a RL agent as a model-free baseline. In contrast to the data-driven model, the RL agent learns to properly perform corrective movements. These results show that it is reasonable to further investigate RL in the context of object pushing. \citeauthor{cong_reinforcement_2022} \cite{cong_reinforcement_2022} therefore proposed a vision-proprioception model for planar object pushing, which is a state-of-the-art RL model that uses latent representations to encode the characteristics of an object. This vision-proprioception model is the starting point of this work. We investigate the following problem: a robot has to precisely push objects with varying physical parameters from starting positions to goal positions, both of which are randomly chosen. The main contribution of this work is the adaption of the vision-proprioception model to push objects more precisely to target positions by using fewer corrective movements around an object. Originally, the task was considered successful if the distance between the center of the object and the goal is smaller than $5\,$cm \cite{cong_reinforcement_2022}, which is a threshold that is commonly used in the literature \cite{plappert_multi-goal_2018, gallouedec_panda-gym_2021, xu_cocoi_2021}. However, this is a comparatively large tolerance, particularly if pushing is considered as an alternative for pick-and-place, which is often used to precisely reposition objects. Therefore, we decrease the threshold to $1\,$cm. The vision-proprioception model is adapted by providing the agent with the complete episode history of observations, using a gated recurrent unit (GRU) \cite{cho_learning_2014} as a feature extractor and improving the sampling of object parameters during training. Fig. \ref{fig_visual_abstract} provides an overview of our approach.
	
	\begin{figure*}[t]
		\centering
		\includegraphics[width=0.98\textwidth]{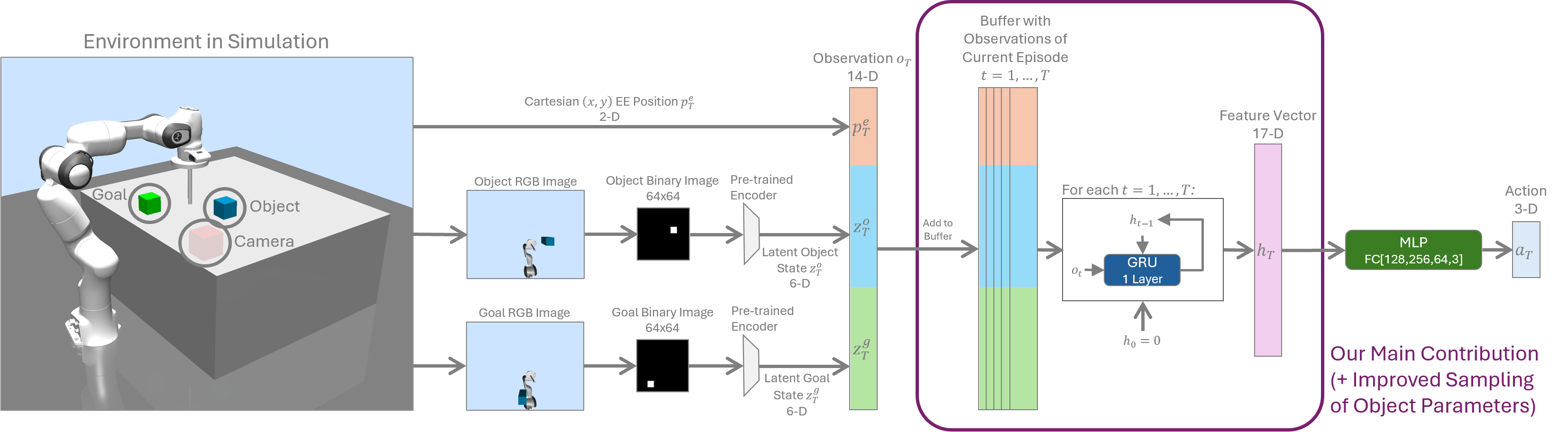}
		\caption{\textbf{Model Architecture.} We concatenate the Cartesian $(x,y)$ EE position with the latent object and goal states generated by an encoder trained prior to the RL agent. The observations of the entire episode are stored in a memory buffer to be processed by a GRU-layer. The hidden state of the most recent time step is used as the feature vector for actor and critic MLPs.}
		\label{fig_model_architecture}
	\end{figure*}
	
	\section{Related Work}
	RL becomes increasingly important in the field of real-world robotic manipulation \cite{braun_grey-box_2023}, \cite{gu_deep_2017}. One important sub class of manipulation is pushing, which has been heavily studied in the past. Pushing is often considered in combination with grasping, to solve a more complex task. More precisely, pushing and grasping can be combined to play the game Jenga \cite{fazeli_see_2019} or in the case of object packing \cite{liang_visuo-tactile_2023}. In other works, pushing is used to singulate objects in cluttered environments \cite{hermans_guided_2012, eitel_learning_2020, kiatos_robust_2019}, e.g. to free space around the object before grasping it \cite{boularias_learning_2015, deng_deep_2019, yang_collaborative_2022, zeng_learning_2018}. Not all approaches comprehensively model pushing actions. Pushing is often an additive skill to rearrange objects in the environment (see \cite{deng_deep_2019}). In all these works, pushing is usually not the main focus. Therefore, the objects mainly differ in their shape and size, but properties such as friction and mass are often not explicitly varied, as most of the work focuses on grasping. However, to precisely push objects to target positions, the sliding friction force is particularly important. Another part of the literature considers object pushing as a benchmark scenario for RL algorithms that focus on multi-goal RL \cite{plappert_multi-goal_2018, gallouedec_panda-gym_2021, andrychowicz_hindsight_2017}, imitation learning \cite{chi_diffusion_2023} or joining multiple input modalities such as vision and touch \cite{chen_visuo-tactile_2022}. In general, these benchmark scenarios are kept simple and object properties, such as shape, size, or friction are usually not changed. Therefore, training an agent that applies to real-world situations requires more sophisticated environments. In addition, there is a lot of work that only focuses on pushing. The approaches range from purely analytical methods to data-driven and deep learning approaches \cite{stuber_lets_2020}. Some more recent methods focus on RL \cite{cong_reinforcement_2022, xu_cocoi_2021, dengler_learning_2022, lowrey_reinforcement_2018}. However, precise pushing is not investigated, as the main focus is usually chosen differently, e.g. on pushing objects through cluttered scenes \cite{dengler_learning_2022} or keeping objects upright while pushing \cite{xu_cocoi_2021}. Moreover, RNNs have already been used in the context of pushing \cite{cong_self-adapting_2020, ajay_augmenting_2018, li_push-net_2018}, but none of these approaches investigate whether using a RNN in combination with RL leads to more precise control of the objects, especially in the case of small sliding friction forces.
	
	\section{Setup}
	\label{sec_setup}
	This work is based on the vision-proprioception model \cite{cong_reinforcement_2022}. In contrast to the original paper, we use a Franka Emika Panda robot with $7$ degrees of freedom and a push rod as the end effector (EE). A camera placed below a glass table provides RGB images of the object and the target. To safely train and test the agent without the risk of damaging hardware, a custom simulation environment is built based on Gymnasium \cite{towers_gymnasium_2023} (formerly OpenAI Gym \cite{brockman_openai_2016}) and the physics engine MuJoCo \cite{todorov_mujoco_2012}. An overview of our setup in simulation is shown in Fig. \ref{fig_model_architecture} (left). Our code is available at: \url{https://github.com/ubi-coro/precise_pushing}
	
	\section{Reinforcement Learning Model}
	\label{sec_rl_basics}
	We consider a finite-horizon Markov decision process (MDP), that is defined by a tuple $\mathcal{M} = (\mathcal{S}, \mathcal{A}, p, r)$, where $\mathcal{S}$ is the continuous state space, $\mathcal{A}$ is the continuous action space, and $p: \mathcal{S} \times \mathcal{A} \times \mathcal{S} \to [0,\infty)$ characterizes the unknown environment dynamics. The task considered in this work belongs to the field of goal-conditioned RL, as objects should be pushed to various goal positions.  Formally, the MDP is augmented by a continuous goal space $\mathcal{G}\subset\mathcal{S}$. At the beginning of an episode, a new goal $g\in\mathcal{G}$ is sampled from a probability distribution over all possible goals with probability density $\rho_{\mathcal{G}}(\cdot)$. The goal does not change within one episode. At each time step $t=1,...,T_{max}$, an observation does not only contain information about the current state of the environment $s_t$ but also about the desired goal $g$, i.e. the agent observes a concatenation of $s_t$ and $g$, see  \cite{schaul_universal_2015}. In addition, the environment emits an immediate reward $r_{t+1}^g := r(s_t, a_t, s_{t+1},g)$ according to the reward function $r:  \mathcal{S} \times \mathcal{A} \times \mathcal{S} \times \mathcal{G} \to \mathbb{R}$. The objective is to find an optimal, parameterized policy $\pi_{\phi,\mathcal{G}}^*$ that maximizes the expected goal-conditioned return:
	\begin{equation}
		\pi_{\phi, \mathcal{G}}^*:= \operatorname*{argmax}_\pi \mathbb{E}_{\pi, \rho_{\mathcal{G}}}\left [\sum_{k=0}^{T_{max}}\gamma^{k} r_{k+1}^g\right]
	\end{equation}
	The subscripts $\pi$ and $\rho_{\mathcal{G}}$ denote that the agent follows the policy $\pi$ and that $g$ is sampled from the probability distribution with probability density $\rho_{\mathcal{G}}(\cdot)$ over all possible goals, respectively.
	
	\subsection{Observation Space}
	An observation contains information about the EE position, the object, and the goal. From the RBG images generated by the camera, 64x64 binary images are obtained through color filtering. At the beginning of an episode, the goal mask is generated by placing the (blue) object at the target position. An autoencoder \cite{rumelhart_learning_1986} is trained prior to the RL agent to reproduce binary images of objects with randomly selected shape, position, and orientation. The latent representations of the object and goal, produced by the encoder, denoted by $z^o, z^g\in\mathbb{R}^6$, are used as encodings of object and goal state. We use a 6-dimensional latent space, as \citeauthor{cong_reinforcement_2022} \cite{cong_reinforcement_2022} report that best results are achieved with this latent space dimension. Therefore, as shown in Fig. \ref{fig_model_architecture}, the current state of the environment $s\in\mathbb{R}^{14}$ is obtained by concatenating the Cartesian planar position of the EE w.r.t. the base frame, denoted by $p^e := [x_e, y_e]^T\in\mathbb{R}^2$, and latent representations $z^o$ and $z^g$.
	
	\subsection{Feature Extraction}
	\label{sec_feat_extraction}
	So far, the observation is similar to the original vision-proprioception model. It is important to note that the agent has no information about the sliding friction force between the object and the table, since the mass and friction coefficients of the object are not visible on the binary images and, therefore, not present in the latent representation $z^o$. However, to precisely reposition an object by pushing, information about the sliding friction force is of great importance. This information can be obtained by observing the distance travelled by the object when a force is applied: the smaller the sliding friction force, the larger the distance travelled. Using the original vision-proprioception model, the agent cannot derive this information, since it does not have access to information from previous observations, i.e. mathematically, the Markov property does not hold. In contrast, we use a GRU-layer \cite{cho_learning_2014} as a feature extractor for actor and critic networks, as shown in Fig. \ref{fig_model_architecture}, to provide the agent with the possibility to extract relevant information about the object's behavior from the complete episode history. We chose a recurrent layer to ensure that the agent is able to handle input sequences of variable length. This is important since the agent can only derive information about the sliding friction force after an initial contact with the object and it should be able to memorize the derived information, as contact with the object is lost in the case of corrective movements. However, the number of time steps needed to approach the initial contact position or to perform a corrective movement can vary greatly depending on the situation. Therefore, variable-length inputs are crucial. Using the GRU feature extractor, the agent observes the complete history of observations of one episode up to the current time step $T\leq T_{max}$. Since the hidden state of a RNN encodes all relevant information from the past and the present, the hidden state of the current time step $T$ is used as the feature vector that is passed to a multi-layer perceptron (MLP). As the recurrent layer is trained together with the actor and critic MLP, the agent learns to decide which information has to be encoded in the hidden state. However, actor and critic networks use separate feature extractors and do not share weights. The initial hidden state of the GRU-layer is set to zero.
	
	\subsection{Action Space}
	\label{sec_action_space}
	An action $a:=[a_x, a_y, a_s]^T$ is continuous and, in contrast to the original vision-proprioception model, three-dimensional, where $a_x,a_y\in[-1,1]$ (unit: m) represent the desired EE position offset in the x and y directions of the base frame. $a_s\in[10,600]$ denotes the number of MuJoCo simulation steps, i.e. the number of robot control cycles, for which to apply the same position offsets $a_x$ and $a_y$. To clarify the meaning of this parameter, we first note that we distinguish between MuJoCo simulation steps and Gymnasium environment steps. At the beginning of a Gymnasium environment step, the agent receives a new observation $s_t$, selects an action $a_t$, and passes a new desired target position of the EE to the controller which remains unchanged throughout the entire period of a single Gymnasium environment step. The time steps $t=1,...,T_{max}$ previously mentioned in section \ref{sec_rl_basics} correspond to these Gymnasium environment steps. The time that passes during one Gymnasium environment step is determined by two parameters. The first parameter is the number $a_s$ of MuJoCo simulation steps that are executed per Gymnasium environment step. The second parameter is the duration of a single MuJoCo simulation step. The latter corresponds to the control cycle of the robot controller and was fixed to $1\,$ms. Thus, the number of MuJoCo simulation steps $a_s$ determines the frequency of environment feedback that the agent receives as well as the time in which the robot controller can converge to the desired goal position. We include this number of simulation steps in the action space to avoid the need for manual tuning, as this parameter is usually crucial for training a RL agent due to the delayed reward problem.
	
	\subsection{Reward Specification}
	We use the following immediate reward function based on the ground truth object and goal position that was shown to yield the best results in \cite{cong_reinforcement_2022}:
	\begin{equation}
		\label{eq_negative_reward_function}
		r(p^o, p^g) := \begin{cases}
			-1,&\text{if }|| p^o - p^g ||_2 \geq 0.01\text{m}\\
			0, &\text{otherwise}
		\end{cases}
	\end{equation}
	where $p^o := [x_o, y_o]^T\in\mathbb{R}^2$ and $p^g := [x_g, y_g]^T\in\mathbb{R}^2$ denote the planar position of the center of object and goal w.r.t. the base frame, respectively. It is important to note that this ground truth information is not part of the observation. The latter only includes the 6-dimensional latent feature vectors extracted by the vision encoder.
	
	\subsection{Episode Termination}
	As indicated by the reward function, an episode is successful if the Euclidean distance between the center of the object and the center of the target is smaller than 1cm in the final time step of an episode. Every episode takes $50$ time steps, i.e. $T_{max}=50$, regardless of whether the goal is reached after fewer time steps. This is important since the object has to remain close to the desired goal position until the end of an episode. 
	
	\section{Reinforcement Learning Algorithms}
	The agent is trained using the Stable-Baselines3 \cite{raffin_stable-baselines3_2021} implementation of Hindsight Experience Replay (HER) \cite{andrychowicz_hindsight_2017} and soft actor-critic (SAC) \cite{haarnoja_soft_2018}. HER increases the sample efficiency by storing additional transitions to the replay buffer. These transitions are generated by changing, i.e. relabeling, the desired goal and recomputing the reward assuming the new goal. To this end, we had to extend the HER implementation to also allow relabeling of non-observation data like the ground truth goal position. Additionally, it is important to note that the GRU feature extractor needs to recalculate all hidden states from $t=1\dotsc T$ based on the new goal encoding. This also required adjustments to the HER implementation of Stable-Baselines3.
	
	\section{Velocity Controller}
	Fig. \ref{fig_model_architecture} shows that the pusher should be perpendicular to the surface of the table to successfully push an object. This constraint is not included in the RL model since it complicates the learning problem, as the agent has to learn both the constraint and the pushing task. Instead, the constraint is solved by a custom velocity controller which moves the EE to a desired (x,y,z) position and keeps the push rod perpendicular to the table.
	
	\section{Domain Randomization}
	We use domain randomization to ensure that the policy can be transferred to a real setup. To this end, the object parameters are sampled at the beginning of each episode from the ranges shown in Table \ref{table_object_params}. All parameters, except the mass and sliding friction coefficient, are sampled uniformly at random. For the mass and the sliding friction, we use a special sampling distribution, introduced in the next section, to focus learning on the difficult scenarios of small friction forces. Additionally, the start and goal positions of an object are randomly chosen at the beginning of each episode. Even if the object’s orientation is not considered in the reward function, the agent should be able to push objects with varying orientations. Since the object and goal are placed on a table, only the initial angle about the z-axis of the local frame has to be sampled. This angle is drawn uniformly at random from the range $[-\pi, \pi]$.
	
	\section{Distribution of the Sliding Friction Force}
	\label{sec_sampling_sliding_fric}
\begin{figure}[]
		\centering
		\def\svgwidth{0.45\textwidth}
		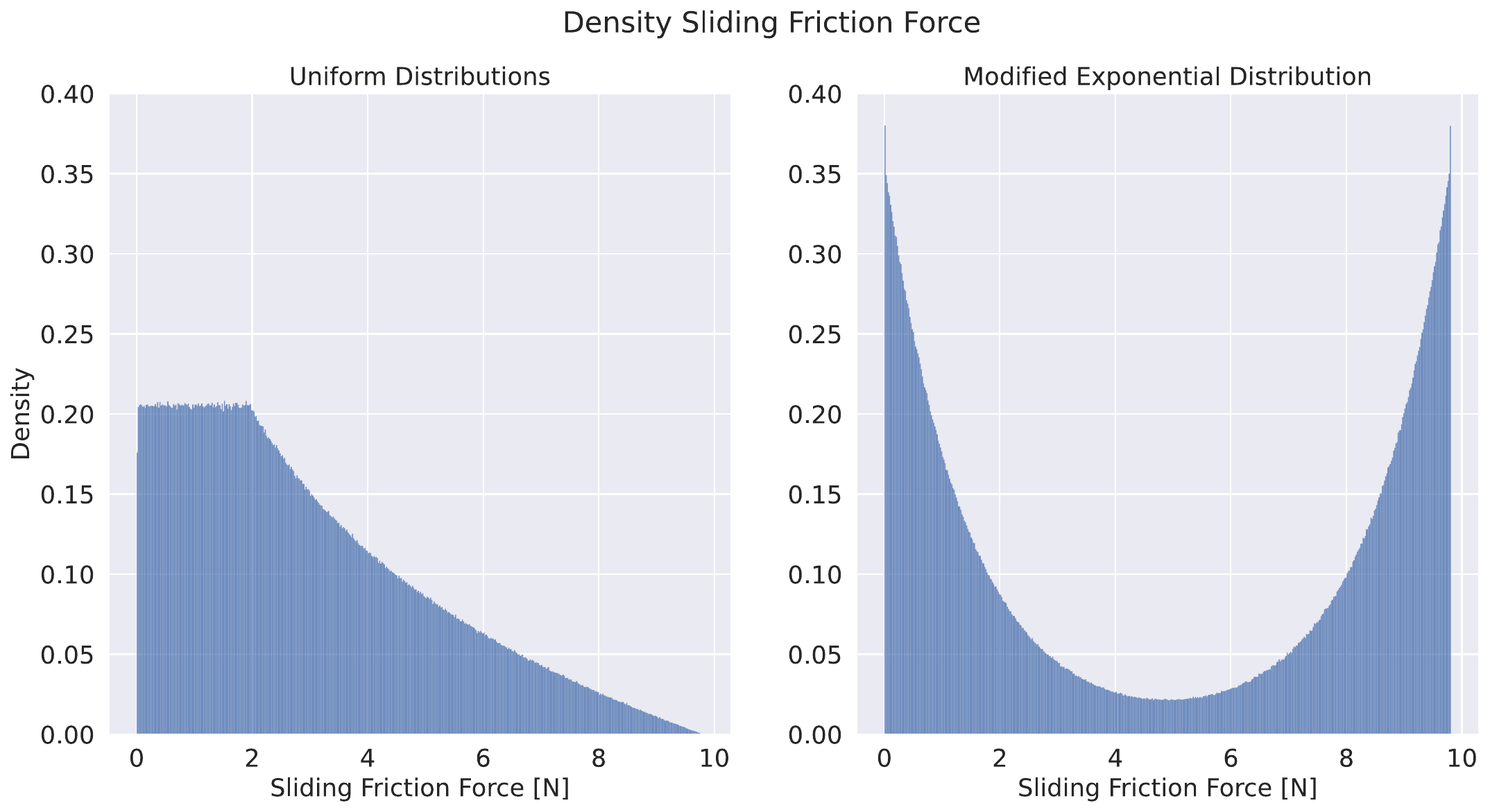
		\caption{\textbf{Densities Sliding Friction Force}. Mass and sliding friction coefficient sampled independently from uniform distributions vs. modified exponential distribution}
		\label{plot_distributions_sliding_friction_force}
	\end{figure}

	As already explained in section \ref{sec_feat_extraction}, the sliding friction force is of great importance when pushing objects. In our setup, the table is not tilted and since MuJoCo uses the maximum friction coefficient to compute the friction forces between any two objects, the sliding friction force $F_k$ between the table and the object is determined by the mass of the object $m_o$ (unit: kg) and the object's sliding friction coefficient $\mu_k$:
	\begin{equation}
		F_k = \mu_k\cdot m_o\cdot9.81\,\mathrm{m}/\mathrm{s^2}
	\end{equation}
	\begin{table}
		\caption{Object Parameter Ranges in Simulation}
		\label{table_object_params}
		\begin{center}
			\begin{tabular}{|l|l|c|l|}
				\hline
				\textbf{Parameter} & \textbf{Range or Value} & \textbf{Unit} \\
				\hline
				Shape & cylinder, cuboid & - \\
				\hline
				Radius cylinder & [0.04, 0.055] & m \\
				\hline
				Length, width cuboid & [0.05, 0.11] & m \\
				\hline
				Minimum height (all shapes) & 0.046 & m \\
				\hline
				Maximum height cuboid & $\mathrm{min}(0.08, \mathrm{length}, \mathrm{width})$ & m \\
				\hline
				Maximum height cylinder & 0.055 & m \\
				\hline
				Mass & [0.001, 1.0] & kg \\
				\hline
				Sliding friction coefficient & [0.2, 1.0] & - \\
				\hline
				Torsional friction coefficient & [0.001, 0.01] & - \\
				\hline
				Damping & 0.01 & - \\
				\hline
				Rolling friction coefficient &  0.0001 & - \\
				\hline
			\end{tabular}
		\end{center}
	\end{table}
	Objects with a small mass and a small sliding friction coefficient, i.e. a small sliding friction force, are the most difficult to reposition precisely. Therefore, we improve the sampling of mass and sliding friction coefficient to ensure that small sliding friction forces are sampled more often during training. More precisely, since mass and sliding friction coefficient equally contribute to the sliding friction force, it is sufficient to sample only the mass value. To this end, minimum and maximum values $\tilde{F}_{k}^{min}$ and $\tilde{F}_{k}^{max}$ are calculated from the ranges of mass, $[m_{o}^{min},m_{o}^{max}]$, and sliding friction coefficient $[\mu_ {k}^{min}, \mu_{k}^{max}]$: 
	\begin{equation}
		\tilde{F}_{k}^{min} = m_{o}^{min}\cdot\mu_{k}^{min}\text{ and }\tilde{F}_{k}^{max} = m_{o}^{max}\cdot\mu_{k}^{max}
	\end{equation}
	Using a fixed friction coefficient $\mu_k=0.4$, the mass has to be sampled from the modified range $\left[\frac{\tilde{F}_{k}^{min}}{\mu_k}, \frac{\tilde{F}_{k}^{max}}{\mu_k}\right]$ to obtain values within the range $[\tilde{F}_{k}^{min},\tilde{F}_{k}^{max}]$. The mass is sampled according to the following distribution:
	\begin{equation}
		m_o = \left(\frac{\tilde{F}_{k}^{max} - \tilde{F}_{k}^{min}}{\mu_k}\right)((1-x)(1-y) +xy)+ \frac{\tilde{F}_{k}^{min}}{\mu_k},
	\end{equation}
	where $x$ is sampled from an exponential distribution with scale parameter $\beta=\frac{1}{7}$ and clipped to the interval $[0,1]$. $y$ is obtained by sampling from a Bernoulli distribution with parameter $p=0.5$. The use of the exponential distribution ensures that the largest values are obtained at the boundaries of the interval, i.e. small or large forces are more likely to be sampled. However, without further modifications, the density of an exponential distribution only decreases which makes it unlikely to sample large forces. Therefore, sampling from the Bernoulli distribution is required to mirror the samples from the exponential distribution by a probability of $50\%$. The resulting density is shown in the right plot of Fig. \ref{plot_distributions_sliding_friction_force}. For comparison, the left plot in Fig. \ref{plot_distributions_sliding_friction_force} shows the probability density of the sliding friction force resulting from the independent uniform sampling of mass and sliding friction coefficient. The plot shows that the density is imbalanced without the improved sampling and that it is comparatively unlikely to sample small and large sliding friction forces during training.
	
	\section{Evaluation}
	\subsection{Experiments in Simulation}
\begin{figure}
		\centering
		\includegraphics[width=0.85\linewidth]{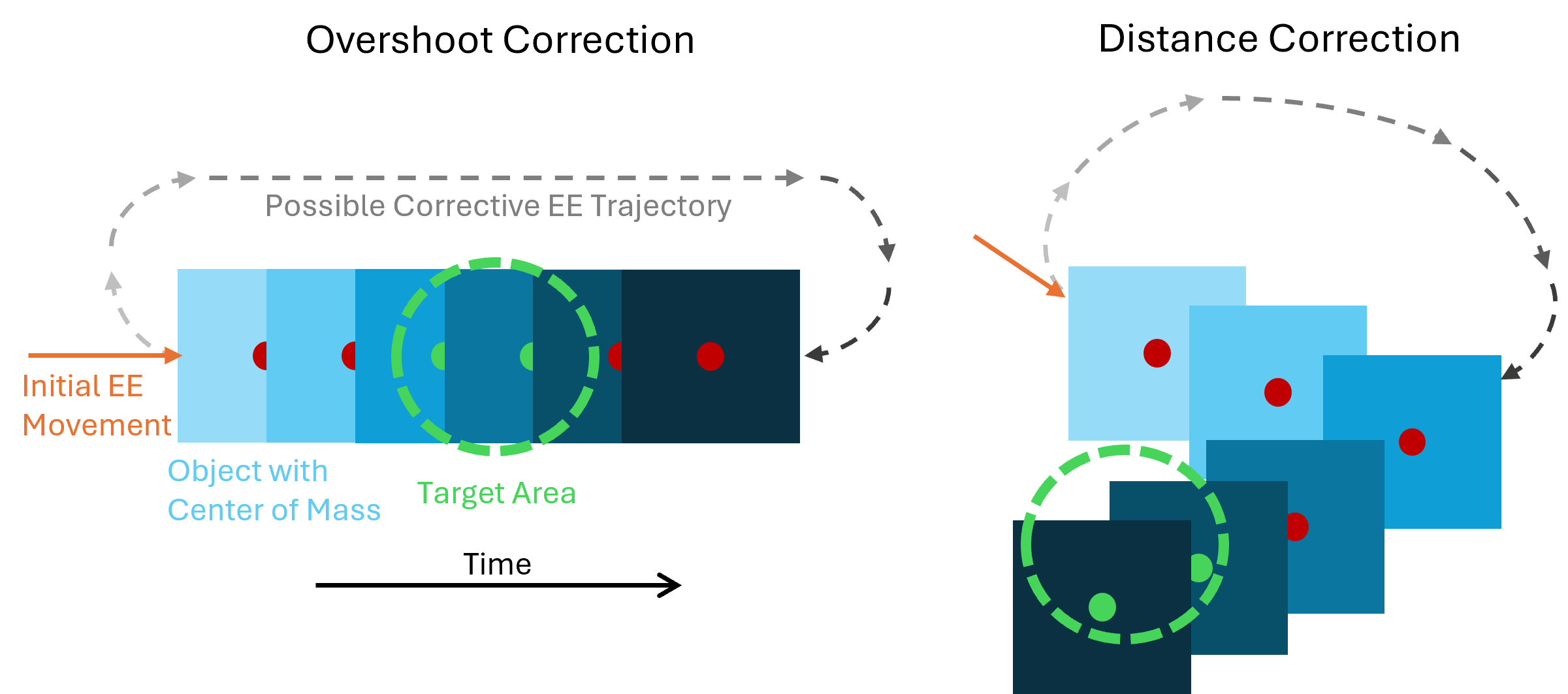}
		\caption{Visualization of the types of corrective movements}
		\label{img_corrective_movements}
	\end{figure}

\begin{figure}[]
		\centering
		\includegraphics[width=\linewidth]{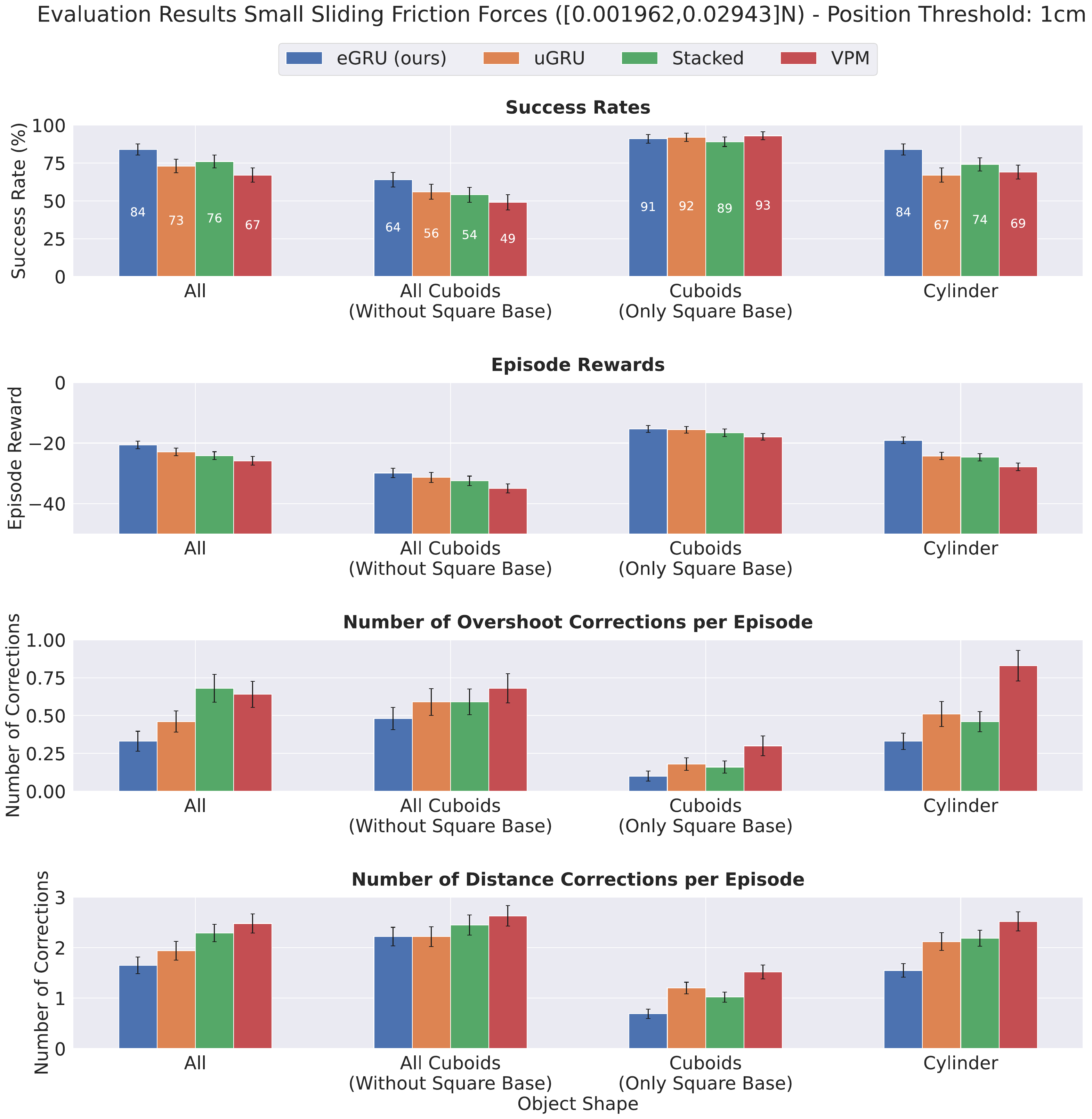}
		\caption{\textbf{Evaluation Results in Simulation.} The results are obtained using a deterministic policy. All object parameters are sampled uniformly at random from the parameter ranges shown in Table \ref{table_object_params}, except the mass (range: $[0.001,0.01]$ kg) and sliding friction coefficient (range: $[0.2,0.3]$). Additionally, we had to adjust the minimum object height for all shapes to $0.052\,$m to avoid distorting the results by disappearing objects due to an unstable simulation.}
		\label{plot_results_smallSF_1cm}
	\end{figure}

	We compare our model (\textit{eGRU}) with the following baselines to show that both main improvements, adding the GRU-layer and improving the sampling of the sliding friction force, are valuable and important for pushing objects precisely:
	\begin{itemize}
		\item \textbf{VPM:} vision-proprioception model from \cite{cong_reinforcement_2022}, but using a smaller position threshold and an action space extended by $a_s$
		\item \textbf{Stacked:} model shown in Fig. \ref{fig_model_architecture}, but using the concatenation of the last 5 observations as a feature vector instead of the hidden state of the GRU-layer
		\item \textbf{uGRU:} model shown in Fig. \ref{fig_model_architecture}
		\item \textbf{eGRU (ours):} model shown in Fig. \ref{fig_model_architecture}, but trained using the improved sampling (see section \ref{sec_sampling_sliding_fric})
	\end{itemize}
	All agents are trained using a position threshold of 1cm and the expanded action space (see section \ref{sec_action_space}). Actor and critic networks of the \textit{VPM} and \textit{Stacked} agent are MLPs with the same architecture as the MLP used for the two GRU agents (see Fig. \ref{fig_model_architecture}). The \textit{VPM}, \textit{Stacked}, and \textit{uGRU} agents are trained without the improved sampling of the sliding friction force, i.e. all object parameters are sampled uniformly at random from the parameter ranges shown in Table \ref{table_object_params}. All agents are evaluated over 100 test episodes. We evaluate all models considering the most difficult case, i.e. objects with a small sliding friction force, by comparing the success rates, the mean episode rewards, and the mean number of corrections within one episode. Two types of corrective movements are distinguished, shown in Fig. \ref{img_corrective_movements}. If the distance between the object and target is smaller than the position threshold in time step $t$ and larger than the position threshold in the next time step $t+1$, a corrective movement is necessary to push the object back to the target area. Therefore, these corrections are referred to as \textit{overshoot corrections}. If the distance between the object and goal increases within one time step and decreases later within the episode, this correction is considered as the second type of corrective movement, i.e. a \textit{distance correction}. In all experiments, the standard error of the mean (SEM) is used to show the variability of the estimated mean.
	The evaluation results depending on the shape of the object are shown in Fig. \ref{plot_results_smallSF_1cm}. The \textit{eGRU} agent (ours) achieves the highest success rates, while the \textit{VPM} model obtains one of the worst success rates. More specifically, the success rates of the \textit{eGRU} agent are about $15\%$ higher than those of the \textit{VPM} agent, except for cuboids with a square base where all agents achieve approximately equal success rates. For all object shape configurations, the \textit{eGRU} agent needs the least number of overshoot and distance corrections, often about half the number of corrective movements compared to the \textit{VPM} agent. In addition, both GRU agents, especially the \textit{eGRU} agent, also obtain larger mean episode rewards indicating that fewer episode time steps are required to successfully push the object to the goal position. This result can be explained by a smaller number of corrective movements usually needed by the GRU agents. The results also show that cuboids with a square base are easiest to control, as all agents achieve the best success rates while receiving the highest episode rewards, but require the fewest corrective movements compared to cuboids with a rectangular base and cylinders. In contrast, cuboids with rectangular bases are the most difficult to control. This is an expected result since the movement of these objects depends on the location of the contact point and whether the contact occurs on the smaller or the longer side of the rectangular base. These results show that our \textit{eGRU} agent clearly outperforms all other baseline agents, especially the original vision-proprioception model.
	
	\subsection{Sim2Real}
\begin{figure}
		\centering
		\includegraphics[width=0.71\linewidth]{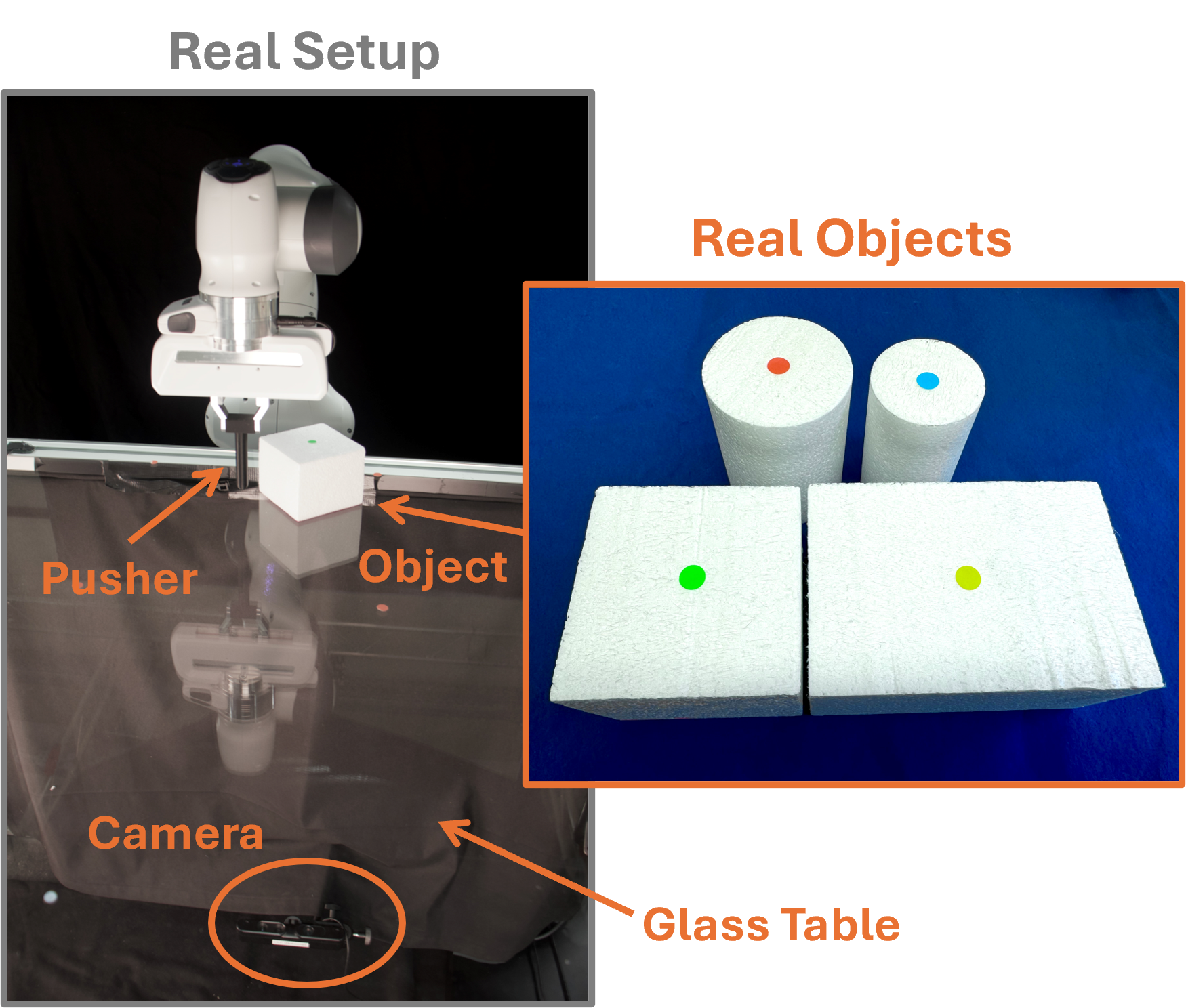}
		\caption{\textbf{Our Real Setup.} We use a Franka Emika Panda robot with a push rod as EE. In addition, we use 4 different objects to test the agents with the real setup.}
		\label{img_real_setup}
	\end{figure}

\begin{figure}
		\centering
		\includegraphics[width=\linewidth]{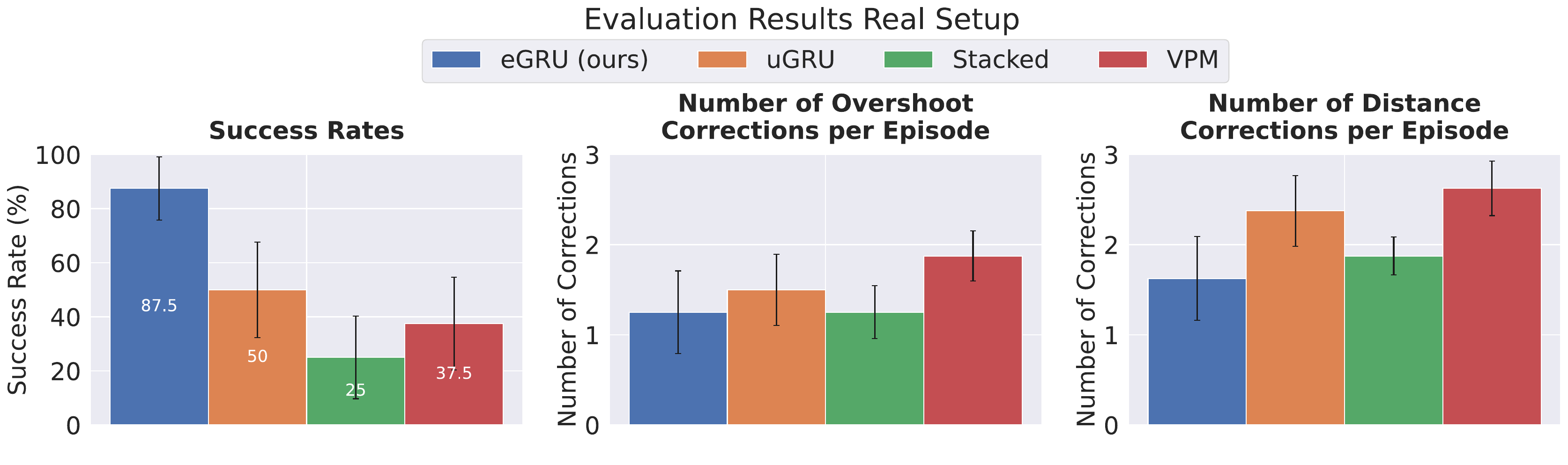}
		\caption{\textbf{Real-World Evaluation Results.} We compare the success rate and the mean number of overshoot and distance corrections per episode.}
		\label{plot_results_real}
	\end{figure}

	Additionally, we evaluate whether our model can be transferred to the real-world setup that is shown in Fig. \ref{img_real_setup}. To this end, we chose four different objects, two of each shape (cylinder and cuboid), as shown in Fig. \ref{img_real_setup}. Masses and sizes of these objects are summarized in Table \ref{table_object_params_real}. The objects are made of polystyrene to make them as light as possible and thus create a challenging task for the agent. All models from the previous section are evaluated over eight episodes, two episodes with each object. The episodes mainly differ in the start and goal positions of the objects. As in the simulation, each episode takes 50 time steps to ensure that the results are comparable. We evaluate the success rate, the mean number of overshoot, and distance corrections. The mean episode reward is not considered, as the ground truth positions are not available in the real setup, such that the reward cannot be computed. For the same reason we judged the success of an episode based on visual inspection only.
	The evaluation results are shown in Fig. \ref{plot_results_real}. Our \textit{eGRU} agent clearly achieves the best success rate at around $90\%$, requiring the least number of both types of corrective movements. The other agents only obtain success rates between $25\%$ and $50\%$, with the \textit{VPM} agent requiring the highest number of corrective movements. When evaluating the agents with our real-world setup, we observed that the \textit{Stacked} agent is often able to push the object to the target position at the beginning of an episode. Thus, the episode would be successful if the agent no longer interacts with the object in the remaining time steps. However, the agent then "corrects" the position of the object, which is not required. This explains the small success rate of the \textit{Stacked} agent. These results show that our \textit{eGRU} agent can be transferred to a real-world setup and that the simulation results can be clearly confirmed. A video of our \textit{eGRU} agent is available at: \url{https://youtu.be/7vD2qtGXxSw}. Interestingly, the agent attempts to achieve not only the goal position of the object but also the goal orientation -- although this task was not explicitly encoded in the reward function.
	
	\begin{table}
		\caption{Masses and Sizes of Real Objects}
		\label{table_object_params_real}
		\begin{center}
			\begin{tabular}{|l|l|l|}
				\hline
				\textbf{Object} & \textbf{Mass [kg]} & \textbf{Size [m]} \\
				\hline
				Blue & 0.006 & Radius: 0.03, Height: 0.1 \\
				\hline
				Red & 0.01 & Radius: 0.04, Height: 0.1 \\
				\hline
				Green & 0.017 & $0.1\times0.1\times0.08$ \\
				\hline
				Yellow & 0.027 & $0.1\times0.15\times0.08$ \\
				\hline
			\end{tabular}
		\end{center}
	\end{table}
	
	\section{Discussion \& Future Work}
	In this paper, we improved the vision-proprioception model, a state-of-the-art RL model, to push objects more precisely to target positions requiring fewer corrective movements. Our main contributions are an improved sampling of the sliding friction force during training and the addition of a GRU-layer to provide the agent with a memory. When applied to a real-world setup, our improved agent achieves a success rate that is about $50\%$ higher and requires approximately $1/3$ fewer corrective movements compared to the original vision-proprioception agent. We show that our model can push objects of two different shapes. However, these shapes are simple and our results clearly show that cuboids with a rectangular base are the most difficult to control. In the future, we plan to investigate imitation learning in the context of object pushing. It will be interesting to evaluate whether such an agent can push objects with more complex shapes. Since the agent would learn from real-world demonstrations, difficult scenarios such as cuboids with a rectangular base might be easier to learn. In addition, the vision-proprioception model assumes that the environment is fully observable, as the camera is placed below a glass table to avoid occlusions. However, a fully observable environment is a rather unrealistic assumption. Therefore, future work should focus on partially observable environments in which the objects might be occluded. Moreover, the policy in this paper provides position offsets, which do not ensure that the obtained motion is smooth. Thus, learning $C^{2}$-continuous control policies should be investigated in future work. Another limitation of this work is the lack of safety guarantees. Tasks that require object pushing are often performed close to people or even in collaboration with humans. Future work should therefore focus on manipulating objects while taking safety constraints into account.
	
	

	
	\section*{ACKNOWLEDGMENT}
	We would like to thank József Lurvig for his contribution to this paper.


	\printbibliography

\end{document}